\documentclass[10pt,twocolumn,letterpaper]{article}

\usepackage{cvpr}
\usepackage{times}
\usepackage{epsfig}
\usepackage{graphicx}
\usepackage{amsmath}
\usepackage{amssymb}
\usepackage{color}
\usepackage{booktabs,threeparttable}
\usepackage{colortbl}
\usepackage[export]{adjustbox}
\usepackage{caption}

\usepackage[pagebackref=true,breaklinks=true,letterpaper=true,colorlinks,bookmarks=false]{hyperref}

\cvprfinalcopy 

\def\cvprPaperID{} 

\begin{document}

\title{You Only Learn One Representation: Unified Network for Multiple Tasks}

\author{Chien-Yao Wang$^{1}$, I-Hau Yeh$^{2}$, and Hong-Yuan Mark Liao$^{1}$ \\
	$^{1}$Institute of Information Science, Academia Sinica, Taiwan\\
	$^{2}$Elan Microelectronics Corporation, Taiwan\\
	{\tt\small kinyiu@iis.sinica.edu.tw, ihyeh@emc.com.tw, and liao@iis.sinica.edu.tw}
}

\maketitle

\begin{abstract}
	People ``understand'' the world via vision, hearing, tactile, and also the past experience. Human experience can be learned through normal learning (we call it explicit knowledge), or subconsciously (we call it implicit knowledge). These experiences learned through normal learning or subconsciously will be encoded and stored in the brain. Using these abundant experience as a huge database, human beings can effectively process data, even they were unseen beforehand. In this paper, we propose a unified network to encode implicit knowledge and explicit knowledge together, just like the human brain can learn knowledge from normal learning as well as subconsciousness learning. The unified network can generate a unified representation to simultaneously serve various tasks. We can perform kernel space alignment, prediction refinement, and multi-task learning in a convolutional neural network. The results demonstrate that when implicit knowledge is introduced into the neural network, it benefits the performance of all tasks. We further analyze the implicit representation learnt from the proposed unified network, and it shows great capability on catching the physical meaning of different tasks. The source code of this work is at : \url{https://github.com/WongKinYiu/yolor}.
\end{abstract}

\section{Introduction}

\begin{figure}[h]
	\begin{center}
		\includegraphics[width=.95\linewidth]{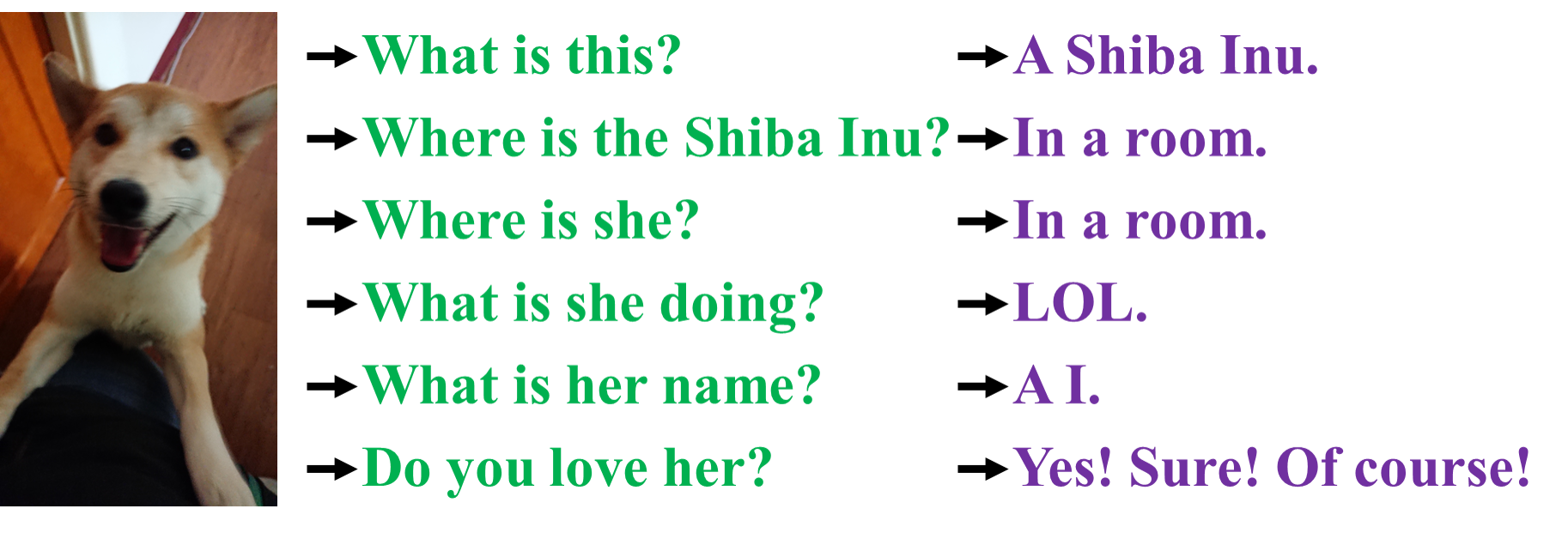}
	\end{center}
	\caption{Human beings can answer different questions from the same input. Our aim is to train a single deep neural network that can serve many tasks.}
	\label{fig:prob}
	\vspace{-6mm}
\end{figure}

As shown in Figure \ref{fig:prob}, humans can analyze the same piece of data from various angles. However, a trained convolutional neural network (CNN) model can usually only fulfill a single objective. Generally speaking, the features that can be extracted from a trained CNN are usually poorly adaptable to other types of problems. The main cause for the above problem is that we only extract features from neurons, and implicit knowledge, which is abundant in CNN, is not used. When the real human brain is operating, the aforementioned implicit knowledge can effectively assist the brain to perform various tasks.

Implicit knowledge refers to the knowledge learned in a subconscious state. However, there is no systematic definition of how implicit learning operates and how to obtain implicit knowledge. In the general definition of neural networks, the features obtained from the shallow layers are often called explicit knowledge, and the features obtained from the deep layers are called implicit knowledge. In this paper, we call the knowledge that directly correspond to observation as explicit knowledge. As for the knowledge that is implicit in the model and has nothing to do with observation, we call it as implicit knowledge.

\begin{figure*}[t]
	\begin{center}
		\includegraphics[width=1.0\linewidth]{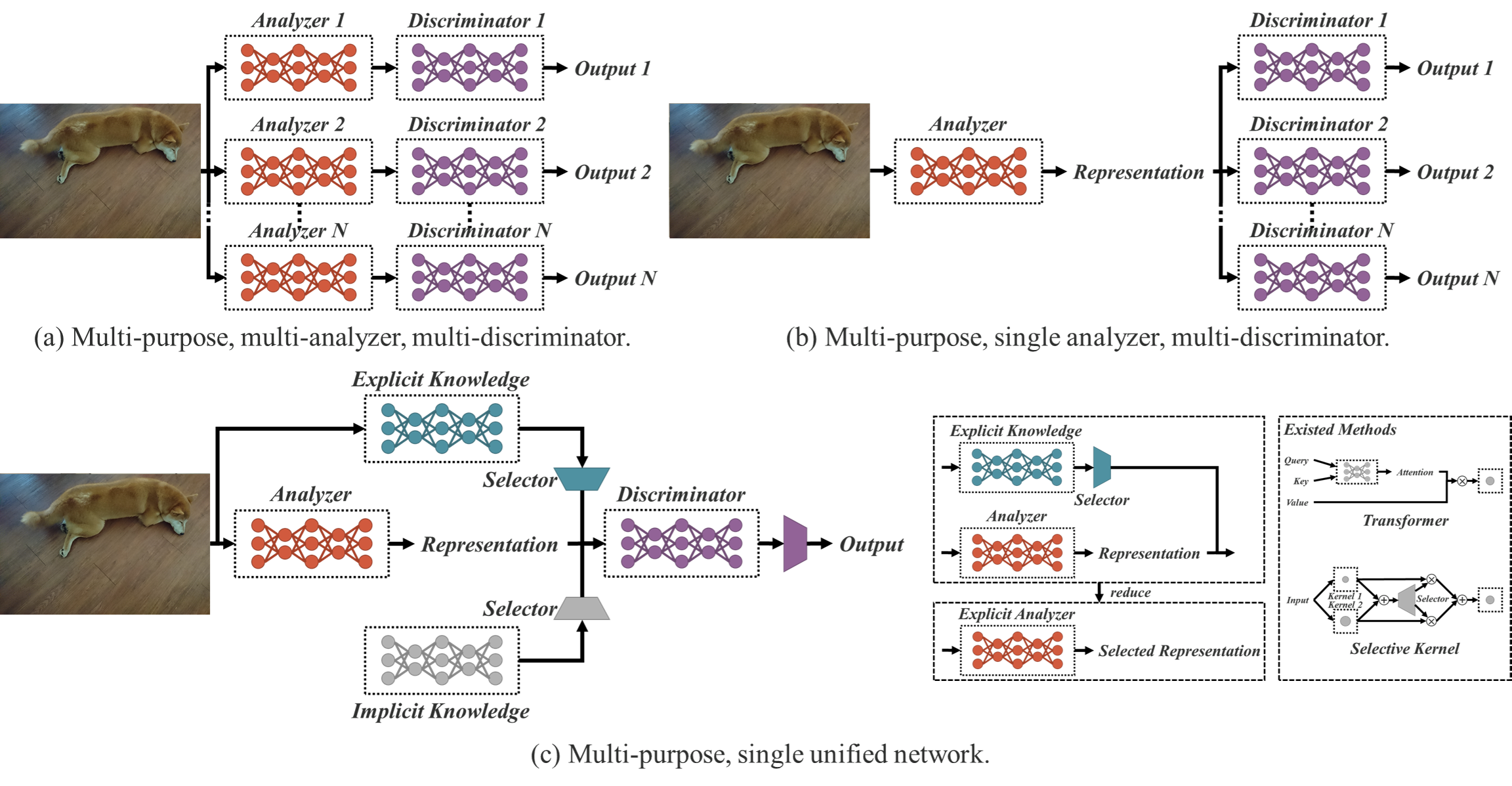}
	\end{center}
	\caption{Multi-purpose NN architectures. (a) distinct models for distinct tasks; (b) shared backbone, different heads for different tasks; and (c) our proposed unified network: one representation with explicit knowledge and implicit knowledge for serving multiple tasks.}
	\vspace{-5mm}
	\label{fig:arch}
\end{figure*}

We propose a unified network to integrate implicit knowledge and explicit knowledge, and enable the learned model to contain a general representation, and this general representation enable sub-representations suitable for various tasks. Figure \ref{fig:arch}.(c) illustrates the proposed unified network architecture.

The way to construct the above unified networks is to combine compressive sensing and deep learning, and the main theoretical basis can be found in our previous work \cite{wang2016robust, wang2017recognition, wang2020sound}. In \cite{wang2016robust}, we prove the effectiveness of reconstructing residual error by extended dictionary. In \cite{wang2017recognition, wang2020sound}, we use sparse coding to reconstruct feature map of a CNN and make it more robust. The contribution of this work are summarized as follows:
\vspace{-2mm}

\begin{enumerate}
	\item We propose a unified network that can accomplish various tasks, it learns a general representation by integrating implicit knowledge and explicit knowledge, and one can complete various tasks through this general representation. The proposed network effectively improves the performance of the model with a very small amount of additional cost (less than one ten thousand of the amount of parameters and calculations.)
	\vspace{-2mm}
	\item We introduced kernel space alignment, prediction refinement, and multi-task learning into the implicit knowledge learning process, and verified their effectiveness.
	\vspace{-2mm}
	\item We respectively discussed the ways of using vector, neural network, or matrix factorization as a tool to model implicit knowledge, and at the same time verified its effectiveness.
	\vspace{-2mm}
	\item We confirmed that the proposed implicit representation learned can accurately correspond to a specific physical characteristic, and we also present it in a visual way. We also confirmed that if operators that conform to the physical meaning of an objective, it can be used to integrate implicit knowledge and explicit knowledge, and it will have a multiplier effect.
	\vspace{-2mm}
	\item Combined with state-of-the-art methods, our proposed unified network achieved comparable accuracy as Scaled-YOLOv4-P7 \cite{wang2020scaled} on object detection and the inference speed has been increased 88\%. 
	\vspace{-2mm}
\end{enumerate}

\section{Related work}

We conduct a review of the literature related to this research topic. This literature review is mainly divided into three aspects: (1) explicit deep learning: it will cover some methods that can automatically adjust or select features based on input data, (2) implicit deep learning: it will cover the related literature of implicit deep knowledge learning and implicit differential derivative, and (3) knowledge modeling: it will list several methods that can be used to integrate implicit knowledge and explicit knowledge.

\subsection{Explicit deep learning}

Explicit deep learning can be carried out in the following ways. Among them, Transformer \cite{vaswani2017attention, carion2020end, wang2021pyramid} is one way, and it mainly uses query, key, or value to obtain self-attention. Non-local networks \cite{wang2018non, cao2019gcnet, yin2020disentangled} is another way to obtain attention, and it mainly extracts pair-wise attention in time and space. Another commonly used explicit deep learning method \cite{li2019selective, zhang2020resnest} is to automatically select the appropriate kernel by input data.

\subsection{Implicit deep learning}

The methods that belong to the category of implicit deep learning are mainly implicit neural representations \cite{sitzmann2020implicit} and deep equilibrium models \cite{bai2019deep, bai2020multiscale, wang2020implicit}. The former is mainly to obtain the parameterized continuous mapping representation of discrete inputs to perform different tasks, while the latter is to transform implicit learning into a residual form neural networks, and perform the equilibrium point calculation on it.

\subsection{Knowledge modeling}

As for the methods belonging to the category of knowledge modeling, sparse representation \cite{aharon2006k, wright2008robust} and memory networks \cite{weston2014memory, sukhbaatar2015end} are mainly included. The former uses exemplar, predefined over complete, or learned dictionary to perform modeling, while the latter relies on combining various forms of embedding to form memory, and enable memory to be dynamically added or changed.

\newpage

\section{How implicit knowledge works?}
\label{sec:how}

The main purpose of this research is to conduct a unified network that can effectively train implicit knowledge, so first we will focus on how to train implicit knowledge and inference it quickly in the follow-up. Since implicit representation $\mathbf{z}_{i}$ is irrelevant to observation, we can think of it as a set of constant tensor $Z = \{\mathbf{z}_{1}, \mathbf{z}_{2}, ..., \mathbf{z}_{k}\}$. In this section we will introduce how implicit knowledge as constant tensor can be applied to various tasks.

\subsection{Manifold space reduction}

\begin{figure}[h]
	\begin{center}
		\includegraphics[width=.95\linewidth]{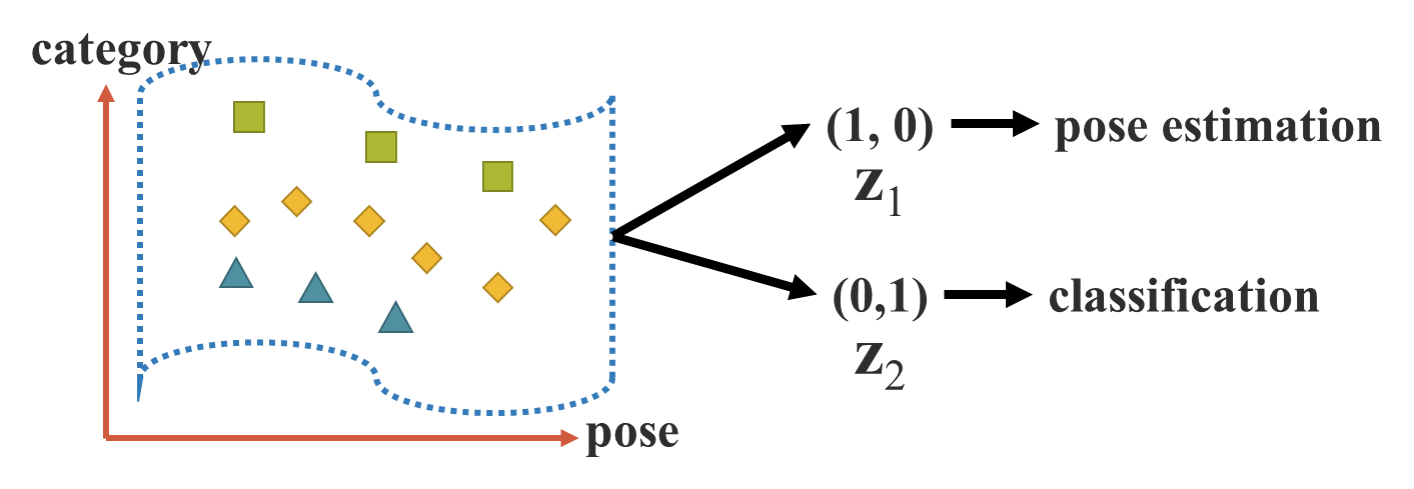}
	\end{center}
	\caption{Manifold space reduction.}
	\label{fig:mani}
\end{figure}

We believe that a good representation should be able to find an appropriate projection in the manifold space to which it belongs, and facilitate the subsequent objective tasks to succeed. For example, as shown in Figure \ref{fig:mani}, if the target categories can be successfully classified by the hyperplane in the projection space, that will be the best outcome. In the above example, we can take the inner product of the projection vector and implicit representation to achieve the goal of reducing the dimensionality of manifold space and effectively achieving various tasks.

\subsection{Kernel space alignment}

\begin{figure}[h]
	\begin{center}
		\includegraphics[width=.75\linewidth]{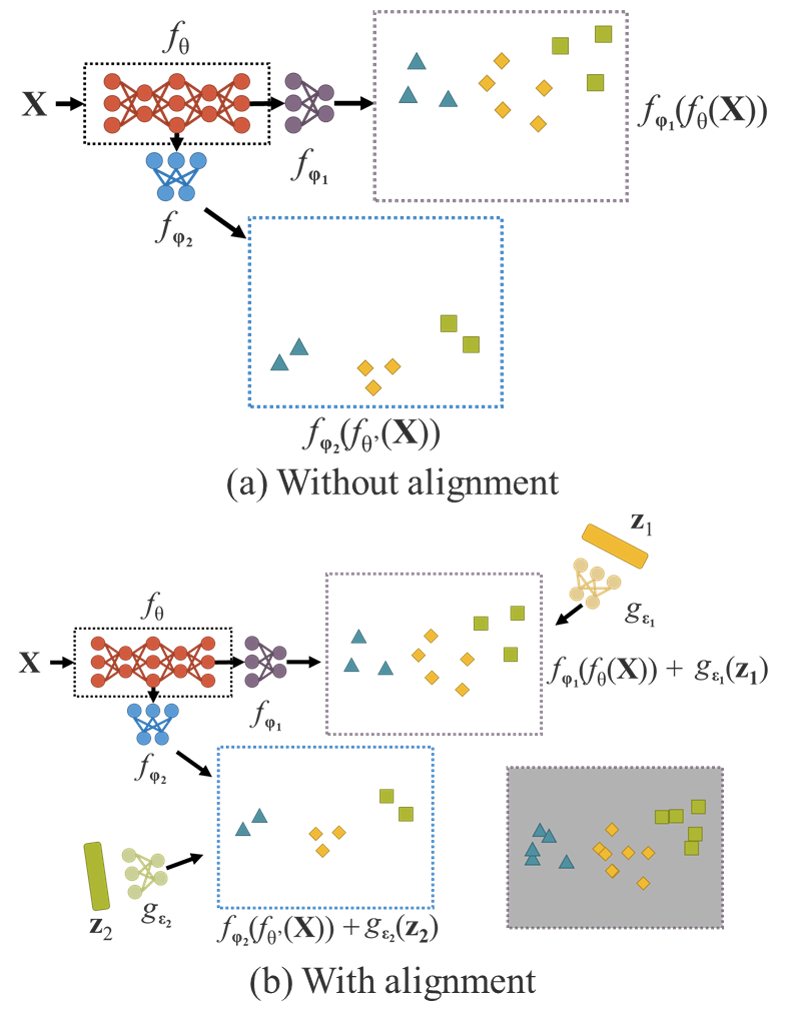}
	\end{center}
	\caption{Kernel space alignment.}
	\label{fig:ker}
\end{figure}

In multi-task and multi-head neural networks, kernel space misalignment is a frequent problem, Figure \ref{fig:ker}.(a) illustrates an example of kernel space misalignment in multi-task and multi-head NN. To deal with this problem, we can perform addition and multiplication of output feature and implicit representation, so that Kernel space can be translated, rotated, and scaled to align each output kernel space of neural networks, as shown in Figure \ref{fig:ker}.(b). The above mode of operation can be widely used in different fields, such as the feature alignment of large objects and small objects in feature pyramid networks (FPN) \cite{lin2017feature}, the use of knowledge distillation to integrate large models and small models, and the handling of zero-shot domain transfer and other issues.

\newpage

\subsection{More functions}

\begin{figure}[h]
	\begin{center}
		\includegraphics[width=.75\linewidth]{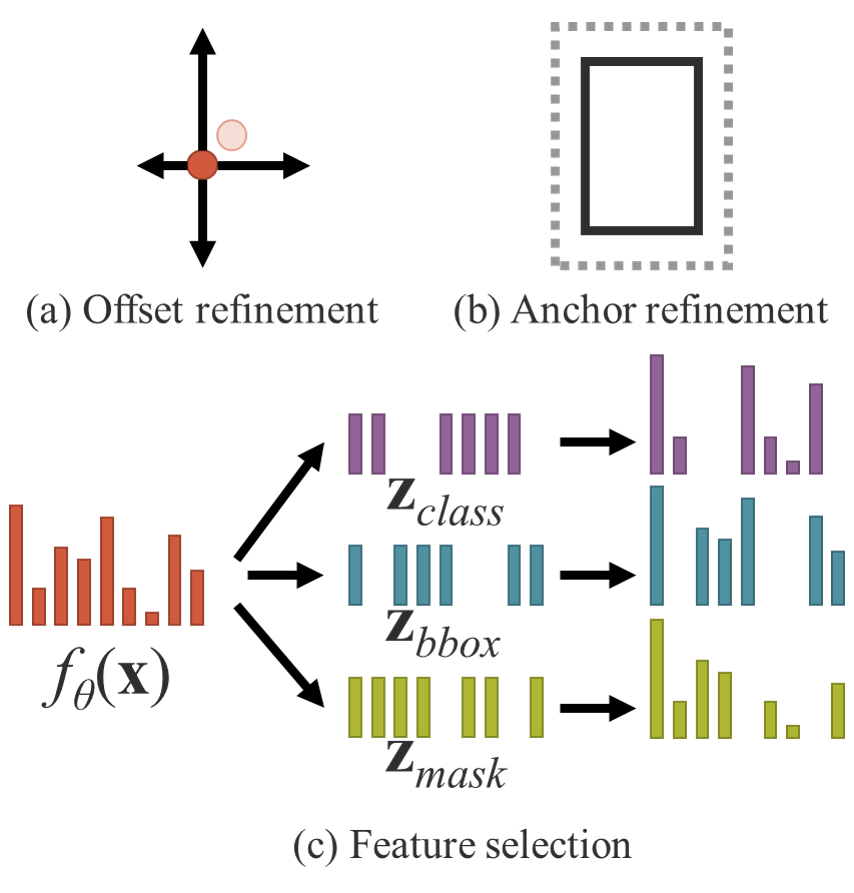}
	\end{center}
	\caption{More functions.}
	\label{fig:func}
\end{figure}

In addition to the functions that can be applied to different tasks, implicit knowledge can also be extended into many more functions.  As illustrated in Figure \ref{fig:func}, through introducing addition, one can make neural networks to predict the offset of center coordinate.  It is also possible to introduce multiplication to automatically search the hyper-parameter set of an anchor, which is very often needed by an anchor-based object detector.  Besides, dot multiplication and concatenation can be used, respectively, to perform multi-task feature selection and to set pre-conditions for subsequent calculations.

\newpage

\section{Implicit knowledge in our unified networks}

In this section, we shall compare the objective function of conventional networks and the proposed unified networks, and to explain why introducing implicit knowledge is important for training a multi-purpose network.  At the same time, we will also elaborate the details of the method proposed in this work.

\subsection{Formulation of implicit knowledge}
\label{sec:fik}

\noindent
\textbf{Conventional Networks:}

For the object function of conventional network training, we can use (\ref{eq:cn}) to express as follows:

\begin{equation}
\begin{split}
& y = f_{\theta}(\mathbf{x}) + \epsilon \\
& \text{minimize } \epsilon
\end{split}
\label{eq:cn}
\end{equation}
where $\mathbf{x}$ is observation, $\theta$ is the set of parameters of a neural network, $f_{\theta}$ represents operation of the neural network, $\epsilon$ is error term, and $y$ is the target of given task.

\begin{figure}[h]
	\vspace{-2mm}
	\begin{center}
		\includegraphics[width=.95\linewidth]{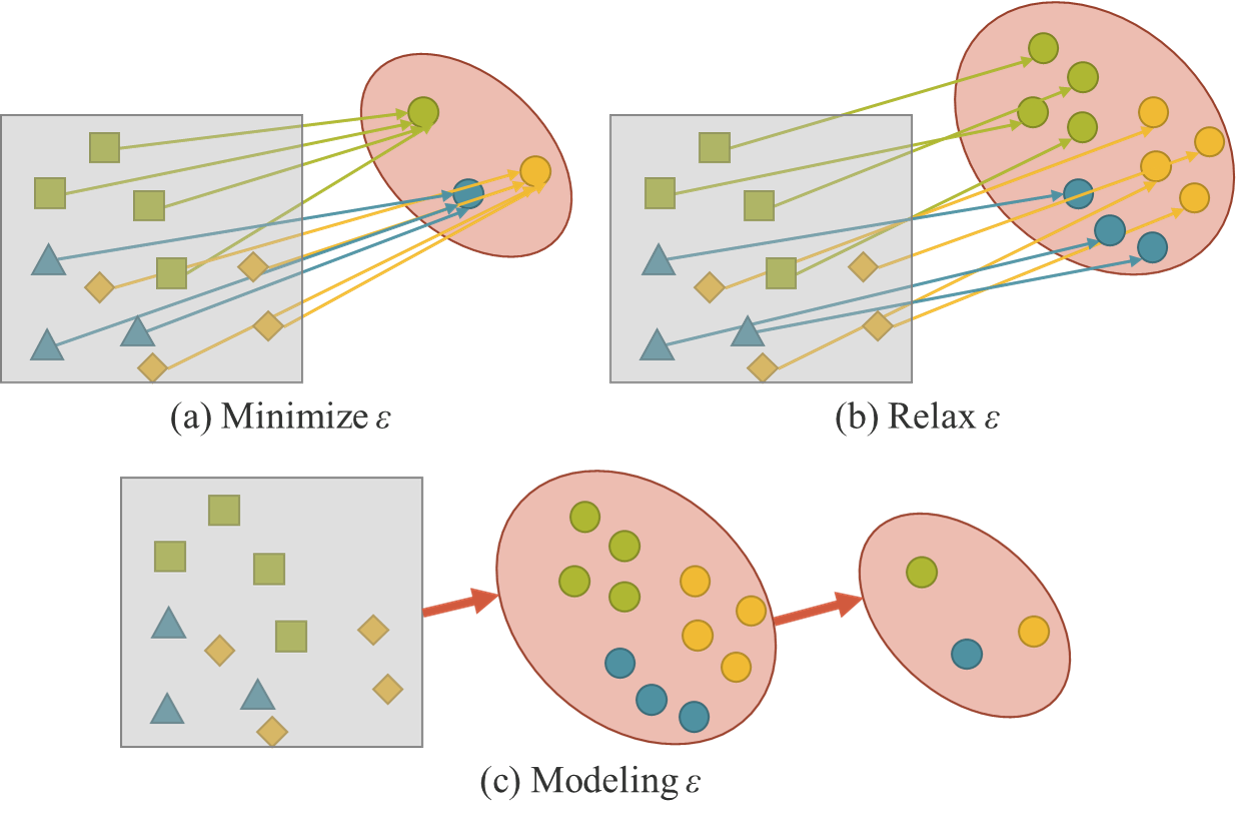}
	\end{center}
	\caption{Modeling error term.}
	\vspace{-4mm}
	\label{fig:spa}
\end{figure}

In the training process of a conventional neural network, usually one will minimize $\epsilon$ to make $f_{\theta}(\mathbf{x})$ as close to the target as possible. This means that we expect different observations with the same target to be a single point in the sub space obtained by $f_{\theta}$, as illustrated in Figure \ref{fig:spa}.(a). In other words, the solution space we expect to obtain is discriminative only for the current task $t_{i}$ and invariant to tasks other than $t_{i}$ in various potential tasks, $T \setminus t_{i}$, where $T = \{t_{1}, t_{2}, ..., t_{n}\}$.

For general purpose neural network, we hope that the obtained representation can serve all tasks belonging to $T$. Therefore, we need to relax $\epsilon$ to make it possible to find solution of each task at the same time on manifold space, as shown in Figure \ref{fig:spa}.(b). However, the above requirements make it impossible for us to use a trivial mathematical method, such as maximum value of one-hot vector, or threshold of Euclidean distance, to get the solution of $t_{i}$. In order to solve the problem, we must model the error term $\epsilon$ to find solutions for different tasks, as shown in Figure \ref{fig:spa}.(c).

\noindent
\textbf{Unified Networks:}

To train the proposed unified networks, we use explicit and implicit knowledge together to model the error term, and then use it to guide the multi-purpose network training process. The corresponding equation for training is as follows:

\begin{equation}
\begin{split}
& y = f_{\theta}(\mathbf{x}) + \epsilon + g_{\phi}(\epsilon_{ex}(\mathbf{x}), \epsilon_{im}(\mathbf{z})) \\
& \text{minimize } \epsilon + g_{\phi}(\epsilon_{ex}(\mathbf{x}), \epsilon_{im}(\mathbf{z}))
\end{split}
\label{eq:un}
\end{equation}
where $\epsilon_{ex}$ and $\epsilon_{im}$ are operations which modeling, respectively, the explicit error and implicit error from observation $\mathbf{x}$ and latent code $\mathbf{z}$. $g_{\phi}$ here is a task specific operation that serves to combine or select information from explicit knowledge and implicit knowledge.

There are some existing methods to integrate explicit knowledge into $f_{\theta}$, so we can rewrite (\ref{eq:un}) into (\ref{eq:ei}).

\begin{equation}
y = f_{\theta}(\mathbf{x}) \star g_{\phi}(\mathbf{z})
\label{eq:ei}
\end{equation}
where $\star$ represents some possible operators that can combine $f_{\theta}$ and $g_{\phi}$. In this work, the operators introduced in Section \ref{sec:how} will be used, which are addition, multiplication, and concatenation.

If we extend derivation process of error term to handling multiple tasks, we can get the following equation:

\begin{equation}
F(\mathbf{x}, \theta, \mathbf{Z}, \Phi, Y, \Psi) = 0
\label{eq:u}
\end{equation}
where $\mathbf{Z} = \{\mathbf{z}_{1}, \mathbf{z}_{2}, ..., \mathbf{z}_{T}\}$ is a set of implicit latent code of $T$ different tasks. $\Phi$ are the parameters that can be used to generate implicit representation from $\mathbf{Z}$. $\Psi$ is used to calculate the final output parameters from different combinations of explicit representation and implicit representation.

For different tasks, we can use the following formula to obtain prediction for all $\mathbf{z} \in \mathbf{Z}$. 

\begin{equation}
d_{\Psi}(f_{\theta}(\mathbf{x}), g_{\Phi}(\mathbf{z}), y) = 0
\label{eq:n}
\end{equation}

For all tasks we start with a common unified representation $f_{\theta}(\mathbf{x})$, go through task-specific implicit representation $g_{\Phi}(\mathbf{z})$, and finally complete different tasks with task-specific discriminator $d_{\Psi}$.

\begin{figure*}[t]
	\begin{center}
		\includegraphics[width=.8\linewidth]{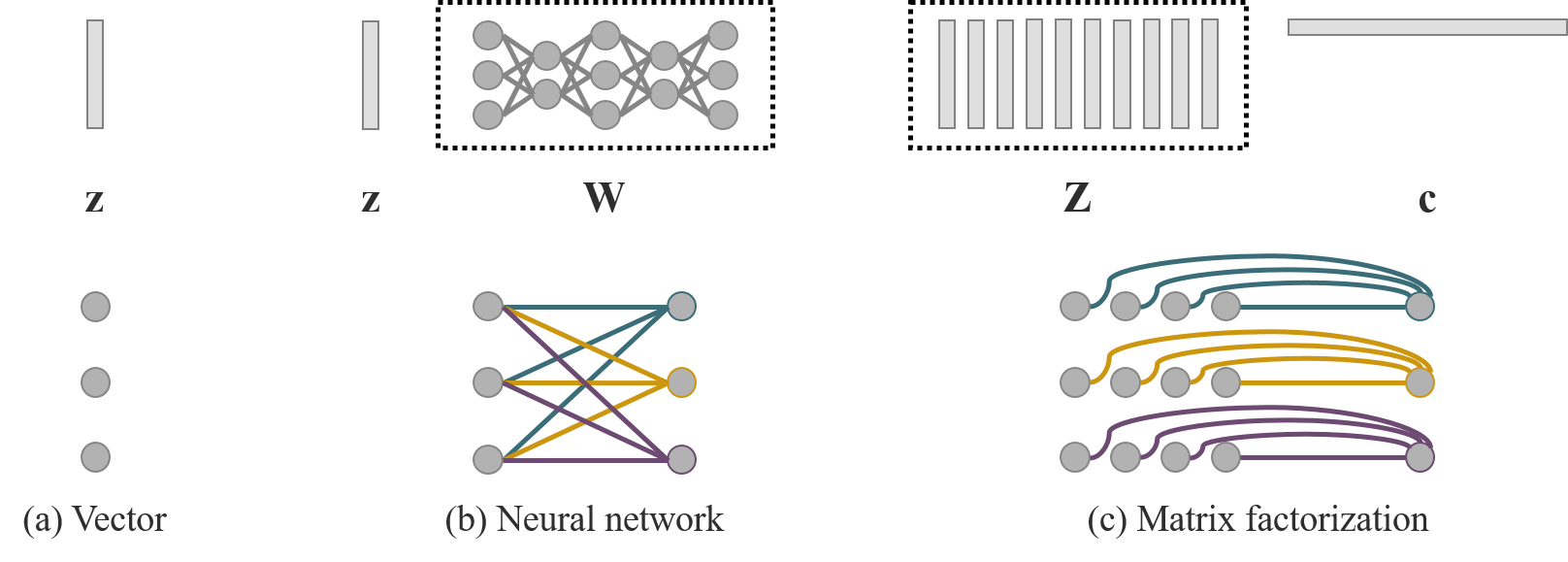}
	\end{center}
	\caption{We proposed to use three different ways for modeling implicit knowledge. The top row shows the formation of these three different modeling approaches, and the bottom row shows their corresponding mathematical attributes. (a) Vector: single base, and each dimension is independent with another dimensions; (b) Neural Network: single or multiple basis, and each dimension is dependent to another dimensions; and (c) Matrix factorization: multiple basis, and each dimension is independent with another dimensions.}
	\label{fig:mod}
\end{figure*}

\newpage

\subsection{Modeling implicit knowledge}
\label{sec:mik}

The implicit knowledge we proposed can be modeled in the following ways:

\noindent
\textbf{Vector / Matrix / Tensor:}

\begin{equation}
\mathbf{z}
\end{equation}

Use vector $\mathbf{z}$ directly as the prior of implicit knowledge, and directly as implicit representation. At this time, it must be assumed that each dimension is independent of each other.

\noindent
\textbf{Neural Network:}

\begin{equation}
\mathbf{Wz}
\end{equation}

Use vector $\mathbf{z}$ as the prior of implicit knowledge, then use the weight matrix $\mathbf{W}$ to perform linear combination or nonlinearization and then become an implicit representation. At this time, it must be assumed that each dimension is dependent of each other. We can also use more complex neural network to generate implicit representation. Or use Markov chain to simulate the correlation of implicit representation between different tasks. 

\noindent
\textbf{Matrix Factorization:}

\begin{equation}
\mathbf{Z^{T}c}
\end{equation}

Use multiple vectors as prior of implicit knowledge, and these implicit prior basis $\mathbf{Z}$ and coefficient $\mathbf{c}$ will form implicit representation. We can also further do sparse constraint to $\mathbf{c}$ and convert it into sparse representation form. In addition, we can also impose non-negative constraint on $\mathbf{Z}$ and $\mathbf{c}$ to convert them into non-negative matrix factorization (NMF) form.

\newpage

\subsection{Training}

Assuming that our model dos not have any prior implicit knowledge at the beginning, that is to say, it will not have any effect on explicit representation $f_{\theta}(\mathbf{x})$. When the combining operator $\star \in \{addition, concatenation\}$, the initial implicit prior  $\mathbf{z} \sim N(0, \sigma)$, and when  the combing operator $\star$ is $multiplication$, $\mathbf{z} \sim N(1, \sigma)$. Here, $\sigma$ is a very small value which is close to zero. As for $\mathbf{z}$ and $\phi$, they both are trained with backpropagation algorithm during the training process.

\subsection{Inference}

Since implicit knowledge is irrelevant to observation $\mathbf{x}$, no matter how complex the implicit model $g_{\phi}$ is, it can be reduced to a set of constant tensor before the inference phase is executed. In other words, the formation of implicit information has almost no effect on the computational complexity of our algorithm. In addition, when the above operator is multiplication, if the subsequent layer is a convolutional layer, then we use (\ref{eq:im}) below to integrate. When one encounters an addition operator, and if the previous layer is a convolutional layer and it has no activation function, then one use (\ref{eq:ia}) shown below to integrate.

 \begin{equation}
 \begin{split}
 \mathbf{x}_{(l+1)} & = \sigma(W_{l}(g_{\phi}(\mathbf{z})\mathbf{x}_{l}) + b_{l}) \\
 & = \sigma(W_{l}^{'}(\mathbf{x}_{l}) + b_{l}), \text{where } W_{l}^{'} = W_{l}g_{\phi}(\mathbf{z})
 \end{split}
 \label{eq:im}
 \end{equation}
 
 \begin{equation}
 \begin{split}
 \mathbf{x}_{(l+1)} & = W_{l}(\mathbf{x}_{l}) + b_{l} +g_{\phi}(\mathbf{z}) \\
 & = W_{l}(\mathbf{x}_{l}) + b_{l}^{'}, \text{where } b_{l}^{'} = b_{l} + g_{\phi}(\mathbf{z})
 \end{split}
 \label{eq:ia}
 \end{equation}
 
 \newpage

\section{Experiments}

Our experiments adopted the MSCOCO dataset \cite{lin2014microsoft}, because it provides ground truth for many different tasks, including \textbf{$^{1}$object detection}, \textbf{$^{2}$instance segmentation}, \textbf{$^{3}$panoptic segmentation}, \textbf{$^{4}$keypoint detection}, \textbf{$^{5}$stuff segmentation}, \textbf{$^{6}$image caption}, \textbf{$^{7}$multi-label image classification}, and \textbf{$^{8}$long tail object recognition}. These data with rich annotation content can help train a unified network that can support computer vision-related tasks as well as natural language processing tasks.

\subsection{Experimental setup}

\begin{figure}[h]
	\begin{center}
		\includegraphics[width=.75\linewidth]{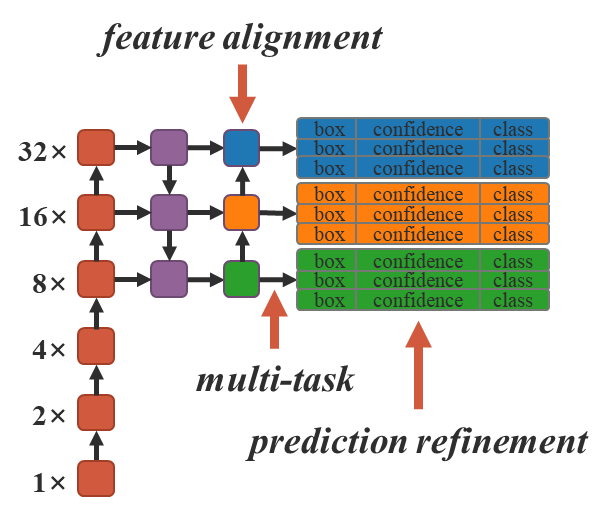}
	\end{center}
	\caption{Architecture.}
	\label{fig:imp}
\end{figure}

In the experimental design, we chose to apply implicit knowledge to three aspects, including \textbf{$^{1}$feature alignment for FPN}, \textbf{$^{2}$prediction refinement}, and \textbf{$^{3}$multi-task learning in a single model}. The tasks covered by multi-task learning include $^{1}$object detection, $^{2}$multi-label image classification, and $^{3}$feature embedding. We choose YOLOv4-CSP \cite{wang2020scaled} as the baseline model in the experiments, and introduce implicit knowledge into the model at the position pointed by the arrow in Figure \ref{fig:imp}. All the training hyper-parameters are compared to default setting of Scaled-YOLOv4 \cite{wang2020scaled}.

In Section \ref{sec:fa}, \ref{sec:pr}, and \ref{sec:mt}, we use the simplist vector implicit representation and addition operator to verify the positive impact on various tasks when implicit knowledge is introduced. In Section \ref{sec:imo}, we will use different operators on different combinations of explicit knowledge and implicit knowledge, and discuss the effectiveness of these combinations. In Section \ref{sec:ima}, we shall model implicit knowledge by using different approaches. In Section \ref{sec:an}, we analyze the model with and without introduce implicit knowledge. Finally in Section \ref{sec:od}, we shall train object detectors with implicit knowledge and then compare the performance with state-of-the-art methods.

\newpage

\subsection{Feature alignment for FPN}
\label{sec:fa}

We add implicit representation into the feature map of each FPN for feature alignment, and the corresponding experiment results are illustrated in Table \ref{table:e1}. From these results shown in Table \ref{table:e1} we can say: After using implicit representation for feature space alignment, all performances, including AP$_{S}$, AP$_{M}$, and AP$_{L}$, have been improved by about 0.5\%, which is a very significant improvement.

\begin{table}[h]
	\centering
	\begin{threeparttable}[h]
		\footnotesize
		\caption{Ablation study of feature alignment.}
		\label{table:e1}
		\setlength\tabcolsep{4.5pt}
		\begin{tabular}{lcccccc}
			\toprule
			\textbf{Model} & \textbf{AP$^{val}$} & \textbf{AP$^{val}_{50}$} & \textbf{AP$^{val}_{75}$} & \textbf{AP$^{val}_{S}$} & \textbf{AP$^{val}_{M}$} & \textbf{AP$^{val}_{L}$} \\				
			\midrule
			\textbf{baseline} & 47.8\% & 66.3\% & 52.1\% & 30.1\% & 52.5\% & 62.0\% \\
			\textbf{+ \textit{i}FA} & \textbf{47.9\%} & \textbf{66.6\%} & \textbf{52.3\%} & \textbf{30.6\%} & \textbf{53.1\%} & \textbf{62.6\%} \\
			\bottomrule
		\end{tabular}
		\begin{tablenotes}[flushleft]
			\footnotesize
			\item[*] baseline is YOLOv4-CSP-fast, tested on 640$\times$640 input resolution.
			\item[*] FA: feature alignment.
		\end{tablenotes}
	\end{threeparttable}
\end{table}

\subsection{Prediction refinement for object detection}
\label{sec:pr}

\begin{figure}[h]
\begin{center}
	\includegraphics[width=.8\linewidth]{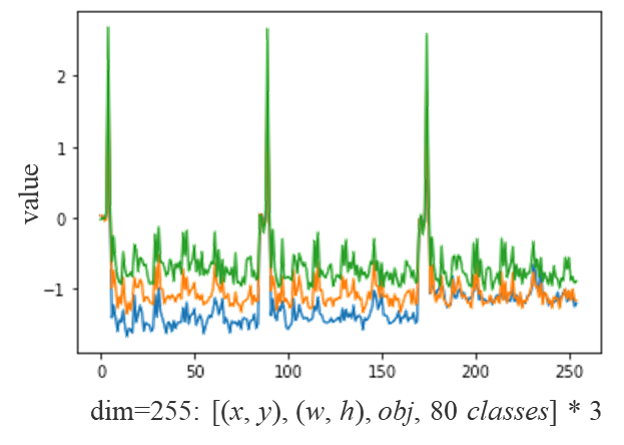}
\end{center}
\caption{Value of learned implicit representation for prediction refinement.}
\label{fig:ipr}
\end{figure}

Implicit representations are added to YOLO output layers for prediction refinement. As illustrated in Table \ref{table:e2}, we see that almost all indicator scores have been improved. Figure \ref{fig:ipr} shows how the introduction of implicit representation affects the detection outcome. In the object detection case, even we do not provide any prior knowledge for implicit representation, the proposed learning mechanism can still automatically learn $(x, y)$, $(w, h)$, $(obj)$, and $(classes)$ patterns of each anchor.

\begin{table}[h]
	\centering
	\begin{threeparttable}[h]
		\footnotesize
		\caption{Ablation study of prediction refinement.}
		\label{table:e2}
		\setlength\tabcolsep{4.5pt}
		\begin{tabular}{lcccccc}
			\toprule
			\textbf{Model} & \textbf{AP$^{val}$} & \textbf{AP$^{val}_{50}$} & \textbf{AP$^{val}_{75}$} & \textbf{AP$^{val}_{S}$} & \textbf{AP$^{val}_{M}$} & \textbf{AP$^{val}_{L}$} \\				
			\midrule
			\textbf{baseline} & 47.8\% & 66.3\% & 52.1\% & 30.1\% & 52.5\% & 62.0\% \\
			\textbf{+ \textit{i}PR} & \textbf{47.8\%} & \textbf{66.5\%} & \textbf{52.1\%} & \textbf{30.3\%} & \textbf{53.3\%} & 61.5\% \\
			\bottomrule
		\end{tabular}
		\begin{tablenotes}[flushleft]
			\footnotesize
			\item[*] baseline is YOLOv4-CSP-fast, tested on 640$\times$640 input resolution.
			\item[*] PR: prediction refinement.
		\end{tablenotes}
	\end{threeparttable}
\end{table}

\newpage

\subsection{Canonical representation for multi-task}
\label{sec:mt}

When one wants to train a model that can be shared by many tasks at the same time, since the joint optimization process on loss function must be executed, multiple parties often pull each other during the execution process. The above situation will cause the final overall performance to be worse than training multiple models individually and then integrating them. In order to solve the above problem, we propose to train a canonical representation for multi-tasks. Our idea is to augment the representation power by introducing implicit representation to each task branch, and the effects it causes are listed in Table \ref{table:e3}. As the data illustrated in Table \ref{table:e3}, without the introduction of implicit representation, some index scores improved after multi-task training, and some dropped. After introducing implicit representation to joint detection and classification (JDC), in the model category corresponding to + \textit{i}JDC, we can clearly see that the overall index score has increased significantly, and it has surpassed the performance of single-task training model. Compared to when implicit representation was not introduced, the performance of our model on medium-sized objects and large-sized objects has also been improved by 0.3\% and 0.7\%, respectively. In the experiment of joint detection and embedding (JDE), because of the characteristic of implicit representation implied by feature alignment, the effect of improving the index score is more significant. Among the index scores corresponding to JDE and + \textit{i}JDE listed in Table \ref{table:e3}, all index scores of + \textit{i}JDE surpass the index that does not introduce implicit representation. Among them, the AP for large objects even increased by 1.1\%.

\begin{table}[h]
	\centering
	\begin{threeparttable}[h]
		\footnotesize
		\caption{Ablation study of multi-task joint learning.}
		\label{table:e3}
		\setlength\tabcolsep{4.5pt}
		\begin{tabular}{lcccccc}
			\toprule
			\textbf{Model} & \textbf{AP$^{val}$} & \textbf{AP$^{val}_{50}$} & \textbf{AP$^{val}_{75}$} & \textbf{AP$^{val}_{S}$} & \textbf{AP$^{val}_{M}$} & \textbf{AP$^{val}_{L}$} \\				
			\midrule
			\textbf{baseline} & 48.0\% & 66.8\% & 52.3\% & 30.0\% & 53.0\% & 62.7\% \\				
			\midrule
			\textbf{JDC} & 47.7\% & \textbf{66.8\%} & 51.9\% & \textbf{30.8\%} & 52.4\% & 61.6\% \\
			\textbf{+ \textit{i}JDC} & \textbf{48.1\%} & \textbf{67.1\%} & 52.2\% & \textbf{31.1\%} & 52.7\% & 62.3\% \\				
			\midrule
			\textbf{JDE} & \textbf{48.1\%} & 66.7\% & \textbf{52.4\%} & \textbf{30.7\%} & \textbf{53.2\%} & 61.9\% \\
			\textbf{+ \textit{i}JDE} & \textbf{48.3\%} & \textbf{66.8\%} & \textbf{52.6\%} & \textbf{30.7\%} & \textbf{53.4\%} & \textbf{63.0\%} \\				
			\bottomrule
		\end{tabular}
		\begin{tablenotes}[flushleft]
		\footnotesize
		\item[*] baseline is YOLOv4-CSP \cite{wang2020scaled}, tested on 640$\times$640 input resolution.
		\item[*] JD\{C, E\}: joint detection \& \{clssification, embedding\}.
	\end{tablenotes}
	\end{threeparttable}
\end{table}

\begin{figure}[h]
\begin{center}
	\includegraphics[width=.95\linewidth]{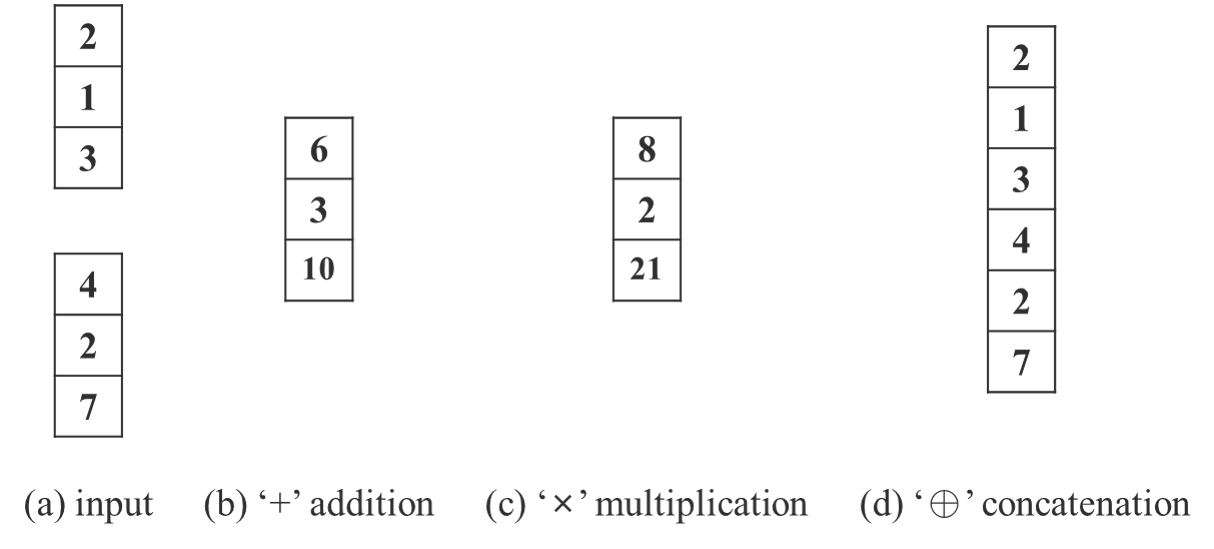}
\end{center}
\caption{Implicit modeling with (a) addition, (b) multiplication, and (c) concatenation operators.}
\label{fig:opr}
\end{figure}

\newpage

\subsection{Implicit modeling with different operators}
\label{sec:imo}

Table \ref{table:e4} shows the experimental results of using different operators shown in Figure \ref{fig:opr} to combine explicit representation and implicit representation. In the implicit knowledge for feature alignment experiment, we see that addition and concatenation both improve performance, while multiplication actually degrades performance. The experimental results of feature alignment are in full compliance with its physical characteristics, because it must deal with the scaling of global shift and all individual clusters. In the implicit knowledge for prediction refinement experiment, since the operator of concatenation ill change the dimension of output, we only compare the effects of using addition and multiplication operators in the experiment. In this set of experiments, the performance of applying multiplication is better than that of applying addition. Analyzing the reason, we found that center shift uses addition decoding when executing prediction, while anchor scale uses multiplication decoding. Because center coordinate is bounded by grid, the impact is minor, and the artificially set anchor owns a larger optimization space, so the improvement is more significant.

\begin{table}[h]
	\centering
	\begin{threeparttable}[h]
		\footnotesize
		\caption{Ablation study of different operators.}
		\label{table:e4}
		\setlength\tabcolsep{4.5pt}
		\begin{tabular}{lcccccc}
			\toprule
			\textbf{Model} & \textbf{AP$^{val}$} & \textbf{AP$^{val}_{50}$} & \textbf{AP$^{val}_{75}$} & \textbf{AP$^{val}_{S}$} & \textbf{AP$^{val}_{M}$} & \textbf{AP$^{val}_{L}$} \\				
			\midrule
			\textbf{baseline} & 47.8\% & 66.3\% & 52.1\% & 30.1\% & 52.5\% & 62.0\% \\				
			\midrule
			\textbf{+ \textit{i}FA} & \textbf{47.9\%} & \textbf{66.6\%} & \textbf{52.3\%} & \textbf{30.6\%} & \textbf{53.1\%} & \textbf{62.6\%} \\
			\textbf{$\times$ \textit{i}FA} & 47.4\% & 65.8\% & 51.6\% & 29.6\% & 52.2\% & \textbf{62.1\%} \\
			\textbf{$\oplus$ \textit{i}FA} & \textbf{47.8\%} & \textbf{66.5\%} & \textbf{52.2\%} & \textbf{30.3\%} & \textbf{52.9\%} & \textbf{62.3\%} \\
			\midrule
			\textbf{+ \textit{i}PR} & \textbf{47.8\%} & \textbf{66.5\%} & \textbf{52.1\%} & \textbf{30.3\%} & \textbf{53.3\%} & 61.5\% \\
			\textbf{$\times$ \textit{i}PR} & \textbf{48.0\%} & \textbf{66.7\%} & \textbf{52.3\%} & 29.8\% & \textbf{53.4\%} & 61.8\% \\
			\bottomrule
		\end{tabular}
		\begin{tablenotes}[flushleft]
			\footnotesize
			\item[*] baseline is YOLOv4-CSP-fast, tested on 640$\times$640 input resolution.
			\item[*] \{+, $\times$, $\oplus$\}: \{addition, multiplication, concatenation\}.
		\end{tablenotes}
	\end{threeparttable}
\end{table}

Based on the above analysis, we designed two other set of experiments -- \{$\times$ \textit{i}FA$^{*}$, $\times$ \textit{i}PR$^{*}$\}. In the first set of experiments -- $\times$ \textit{i}FA$^{*}$, we split feature space into anchor cluster level for combination with multiplication, while in the second set of experiments -- $\times$ \textit{i}PR$^{*}$, we only performed multiplication refinement on width and height in prediction. The results of the above experiments are illustrated in Table \ref{table:e5}. From the figures shown in Table \ref{table:e5}, we find that after corresponding modifications, the scores of various indices have been comprehensively improved. The experiment shows that when we designing how to combine explicit and implicit knowledge, we must first consider the physical meaning of the combined layers to achieve a multiplier effect.

\begin{table}[h]
	\vspace{-2mm}
	\centering
	\begin{threeparttable}[h]
		\footnotesize
		\caption{Ablation study of different operators.}
		\label{table:e5}
		\setlength\tabcolsep{4.5pt}
		\begin{tabular}{lcccccc}
			\toprule
			\textbf{Model} & \textbf{AP$^{val}$} & \textbf{AP$^{val}_{50}$} & \textbf{AP$^{val}_{75}$} & \textbf{AP$^{val}_{S}$} & \textbf{AP$^{val}_{M}$} & \textbf{AP$^{val}_{L}$} \\				
			\midrule
			\textbf{baseline} & 47.8\% & 66.3\% & 52.1\% & 30.1\% & 52.5\% & 62.0\% \\				
			\midrule
			\textbf{$\times$ \textit{i}FA$^{*}$} & \textbf{47.9\%} & \textbf{66.6\%} & 52.0\% & \textbf{30.5\%} & \textbf{52.6\%} & \textbf{62.3\%} \\
			\textbf{$\times$ \textit{i}PR$^{*}$} & \textbf{48.1\%} & \textbf{66.5\%} & \textbf{52.1\%} & \textbf{30.1\%} & \textbf{53.3\%} & 61.9\% \\
			\bottomrule
		\end{tabular}
		\begin{tablenotes}[flushleft]
			\footnotesize
			\item[*] baseline is YOLOv4-CSP-fast, tested on 640$\times$640 input resolution.
		\end{tablenotes}
	\end{threeparttable}
    \vspace{-2mm}
\end{table}

\newpage

\subsection{Modeling implicit knowledge in different ways}
\label{sec:ima}

We tried to model implicit knowledge in different ways, including vector, neural networks, and matrix factorization. When modeling with neural networks and matrix factorization, the default value of implicit prior dimension is twice that of explicit representation dimension. The results of this set of experiments are shown in Table \ref{table:e6}. We can see that whether it is to use neural networks or matrix factorization to model implicit knowledge, it will improve the overall effect. Among them, the best results have been achieved by using matrix factorization model, and it upgrades the performance of AP, AP$_{50}$, and AP$_{75}$ by 0.2\%, 0.4\%, and 0.5\%, respectively. In this experiment, we demonstrated the effect of using different modeling ways. Meanwhile, we confirmed the potential of implicit representation in the future.

\begin{table}[h]
	\centering
	\vspace{-2mm}
	\begin{threeparttable}[h]
		\footnotesize
		\caption{Ablation study of different modeling approaches.}
		\label{table:e6}
		\setlength\tabcolsep{4.5pt}
		\begin{tabular}{lcccccc}
			\toprule
			\textbf{Model} & \textbf{AP$^{val}$} & \textbf{AP$^{val}_{50}$} & \textbf{AP$^{val}_{75}$} & \textbf{AP$^{val}_{S}$} & \textbf{AP$^{val}_{M}$} & \textbf{AP$^{val}_{L}$} \\				
			\midrule
			\textbf{baseline} & 47.8\% & 66.3\% & 52.1\% & 30.1\% & 52.5\% & 62.0\% \\				
			\midrule
			\textbf{+ \textit{i}FA} & \textbf{47.9\%} & \textbf{66.6\%} & \textbf{52.3\%} & \textbf{30.6\%} & \textbf{53.1\%} & \textbf{62.6\%} \\
			\textbf{+ \textit{wi}FA} & \textbf{47.8\%} & \textbf{66.4\%} & 52.0\% & \textbf{30.8\%} & \textbf{52.8\%} & 61.9\% \\
			\textbf{+ \textit{ic}FA} & \textbf{48.0\%} & \textbf{66.7\%} & \textbf{52.6\%} & \textbf{30.3\%} & \textbf{53.2\%} & \textbf{62.5\%} \\
			\bottomrule
		\end{tabular}
		\begin{tablenotes}[flushleft]
			\footnotesize
			\item[*] baseline is YOLOv4-CSP-fast, tested on 640$\times$640 input resolution.
			\item[*] \{\textit{i}, \textit{wi}, \textit{ic}\}: \{vector, neural network, matrix factorization\}, see \ref{sec:mik}.
		\end{tablenotes}
	\end{threeparttable}
    \vspace{-4mm}
\end{table}

\subsection{Analysis of implicit models}
\label{sec:an}

\begin{figure}[h]
	\vspace{-4mm}
	\begin{center}
		\includegraphics[width=.85\linewidth]{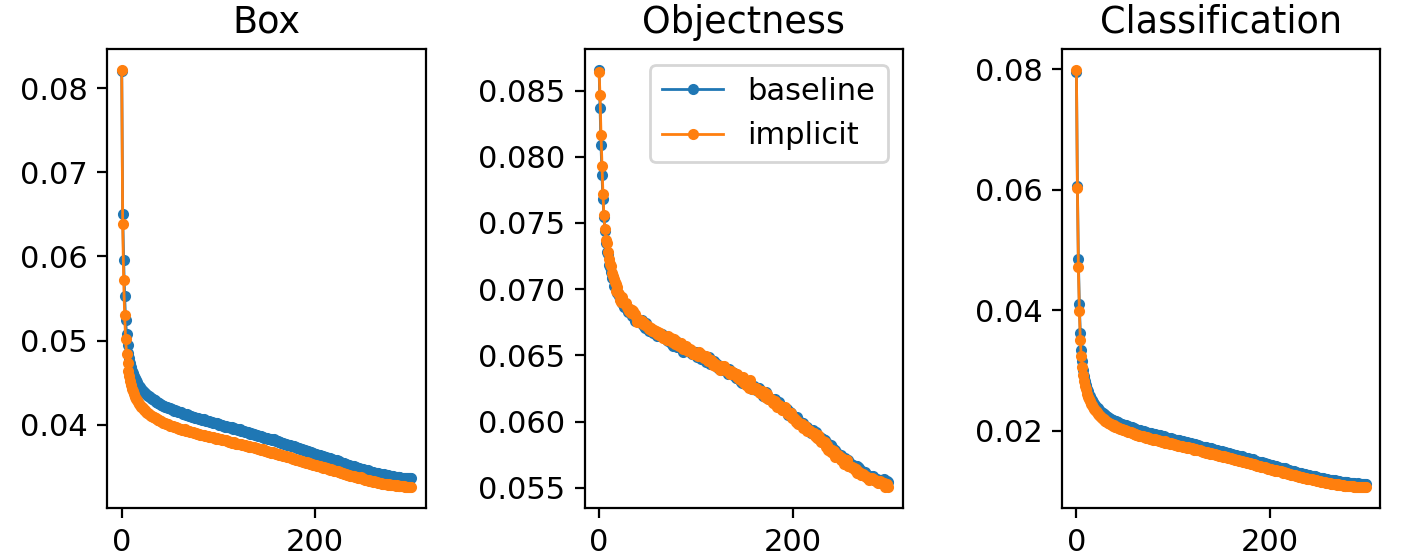}
	\end{center}
	\vspace{2mm}
	\caption{Learning curve of model with/w/o implicit knowledge.}
	\label{fig:con}
	\vspace{-2mm}
\end{figure}

We analyze the number of parameters, FLOPs, and learning process of model with/w/o implicit knowledge, and show the results in Table \ref{table:pf} and Figure \ref{fig:con}, respectively. From the experimental data, we found that in the model with implicit knowledge set of experiments, we only increased the amount of parameters and calculations by less than one ten thousandth, which can significantly improve the performance of the model, and the training process can also converge quickly and correctly.

\begin{table}[h]
	\vspace{-2mm}
	\centering
	\begin{threeparttable}[h]
		\footnotesize
		\caption{Information of model with/without implicit knowledge.}
		\label{table:pf}
		\setlength\tabcolsep{3.5pt}
		\begin{tabular}{lccc}
			\toprule
			\textbf{Model} & \textbf{AP$^{val}$} & \textbf{\# parameters} & \textbf{MFLOPs} \\				
			\midrule
			\textbf{baseline 1} & 47.8\% & 52908989 & 117517.2952 \\
			\textbf{implicit 1} & \textbf{48.0\%} & \textbf{52911546} (+0.005\%) & \textbf{117519.4372} (+0.002\%) \\	
			\midrule
			\textbf{baseline 2} & 51.4\% & 37262204 & 326256.1624 \\
			\textbf{implicit 2} & \textbf{51.9\%} & \textbf{37265016} (+0.008\%) & \textbf{326264.7304} (+0.003\%) \\		
			\bottomrule
		\end{tabular}
		\begin{tablenotes}[flushleft]
			\footnotesize
			\item[*] baseline 1 is YOLOv4-CSP-fast, tested on 640$\times$640 input resolution.
			\item[*] baseline 2 is YOLOv4-P6-light, tested on 1280$\times$1280 input resolution.
			\item[*] implicit \{1, 2\} are baseline \{1, 2\} with + \textit{i}FA, $\times$ \textit{i}PR.
		\end{tablenotes}
	\end{threeparttable}
    \vspace{-2mm}
\end{table}

\newpage

\subsection{Implicit knowledge for object detection}
\label{sec:od}

Finally, we compare the effectiveness of the proposed method with object detection's state-of-the-art methods. The benefits of introducing implicit knowledge are shown in Table \ref{table:as}. For the entire training process, we follow the scaled-YOLOv4 \cite{wang2020scaled} training process, that is, train from scratch 300 epochs first, and then fine-tune 150 epochs. Table \ref{table:sota} illustrates the comparisons with the state-of-the-art methods. One thing worth noting is that our proposed method does not have additional training data and annotations. By introducing the unified network of implicit knowledge, we still achieve results that are sufficient to match the state-of-the-art methods.

\begin{table}[h]
	\centering
	\vspace{-2mm}
	\begin{threeparttable}[h]
		\footnotesize
		\caption{Benefit from implicit knowledge.}
		\label{table:as}
		\setlength\tabcolsep{2.5pt}
		\begin{tabular}{lcccccc}
			\toprule
			\textbf{Model} & \textbf{AP$^{val}$} & \textbf{AP$^{val}_{50}$} & \textbf{AP$^{val}_{75}$} & \textbf{AP$^{val}_{S}$} & \textbf{AP$^{val}_{M}$} & \textbf{AP$^{val}_{L}$} \\				
			\midrule
			\textbf{baseline} & 51.4\% & 69.5\% & 56.4\% & 35.2\% & 55.8\% & 64.6\% \\
			\textbf{implicit} & \textbf{51.9\%} & \textbf{69.8\%} & \textbf{56.8\%} & \textbf{36.0\%} & \textbf{56.3\%} & \textbf{65.0\%} \\
			\textbf{fine-tuned implicit} & \textbf{52.5\%} & \textbf{70.5\%} & \textbf{57.6\%} & \textbf{37.1\%} & \textbf{57.2\%} & \textbf{65.4\%} \\		
			\bottomrule
		\end{tabular}
		\begin{tablenotes}[flushleft]
			\footnotesize
			\item[*] baseline is YOLOv4-P6-light, tested on 1280$\times$1280 input resolution.
			\item[*] implicit is baseline with + \textit{i}FA, $\times$ \textit{i}PR.
		\end{tablenotes}
	\end{threeparttable}
    \vspace{-4mm}
\end{table}

\begin{table}[h]
\centering
\vspace{-2mm}
\begin{threeparttable}[h]
	\footnotesize
	\caption{Comparion of state-of-the-art.}
	\label{table:sota}
	\setlength\tabcolsep{.8pt}
	\begin{tabular}{lccccccc}
		\toprule
		\textbf{Method} & \textbf{pre.} & \textbf{seg.} & \textbf{add.} & \textbf{AP$^{test}$} & \textbf{AP$^{test}_{50}$} & \textbf{AP$^{test}_{75}$} & \textbf{FPS$^{V100}$} \\				
		\midrule
		\textbf{YOLOR (ours)} &  &  &  & 55.4\% & 73.3\% & 60.6\% & 30 \\
		\textbf{ScaledYOLOv4 \cite{wang2020scaled}} &  &  &  & 55.5\% & 73.4\% & 60.8\% & 16 \\	
		\textbf{EfficientDet \cite{tan2019efficientdet}} & $\checkmark$ &  &  & 55.1\% & 74.3\% & 59.9\% & 6.5 \\	
		\textbf{SwinTransformer \cite{liu2021swin}} & $\checkmark$ & $\checkmark$ &  & 57.7\% & -- & -- & -- \\	
		\textbf{CenterNet2 \cite{zhou2021probabilistic}} & $\checkmark$ &  & $\checkmark$ & 56.4\% & 74.0\% & 61.6\% & -- \\	
		\textbf{CopyPaste \cite{ghiasi2020simple}} & $\checkmark$ & $\checkmark$ & $\checkmark$ & 57.3\% & -- & -- & -- \\
		\bottomrule
	\end{tabular}
	\begin{tablenotes}[flushleft]
		\footnotesize
		\item[*] pre. : large dataset image classification pre-training.
		\item[*] seg. : training with segmentation ground truth.
		\item[*] add. : training with additional images.
	\end{tablenotes}
\end{threeparttable}
\vspace{-6mm}
\end{table}

\begin{figure}[t]
	\begin{center}
		\includegraphics[width=.85\linewidth]{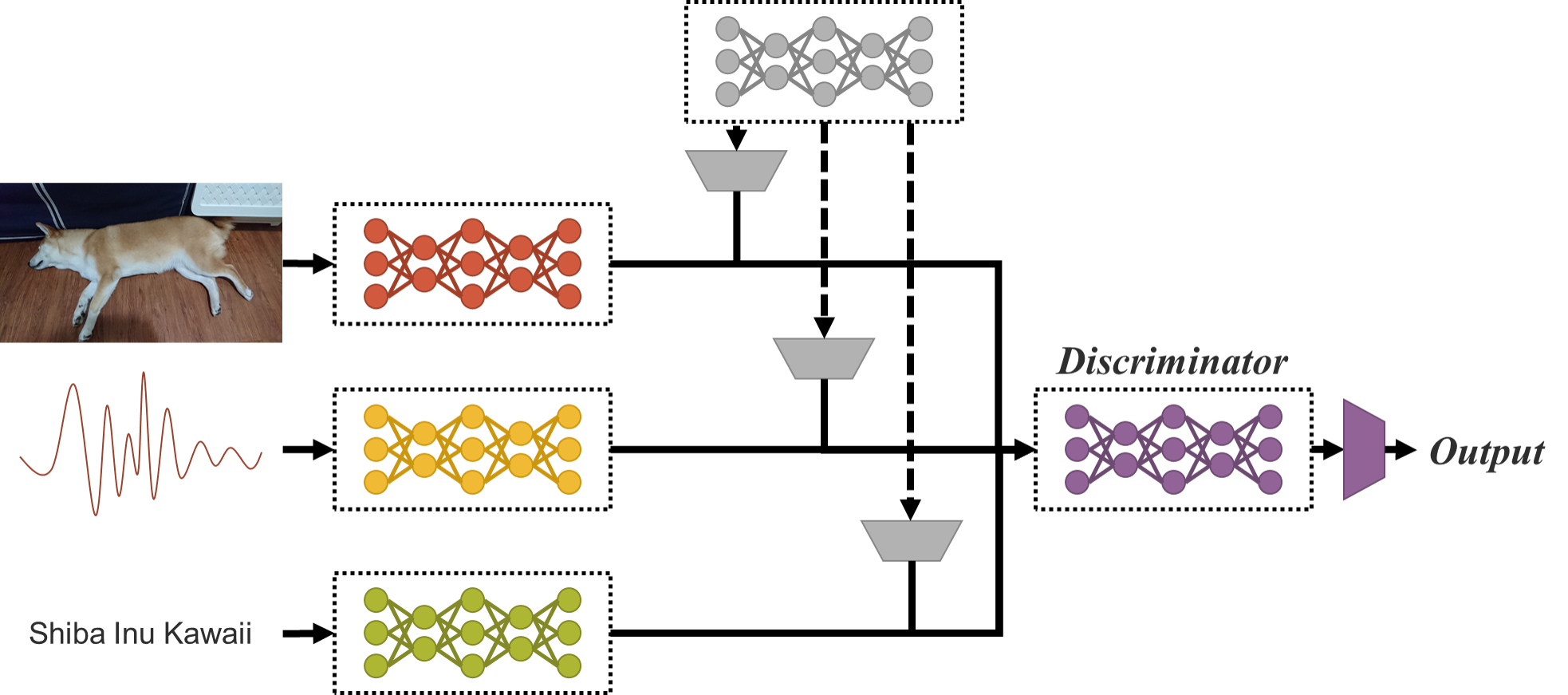}
	\end{center}
    \vspace{2mm}
	\caption{Multimodal unified netwrok.}
	\label{fig:fut}
	\vspace{-6mm}
\end{figure}

\section{Conclusions}

In this paper, we show how to construct a unified network that integrates implicit knowledge and explicit knowledge, and prove that it is still very effective for multi-task learning under the single model architecture. In the future, we shall extend the training to multi-modal and multi-task, as shown in Figure \ref{fig:fut}.

\section{Acknowledgements}

The authors wish to thank National Center for High-performance Computing (NCHC) for
providing computational and storage resources.


\newpage

\appendix

\setcounter{table}{0}
\renewcommand{\thetable}{A\arabic{table}}

\setcounter{figure}{0}
\renewcommand{\thefigure}{A\arabic{figure}}

\section{Appendix}

\begin{figure}[h]
	\begin{center}
		\includegraphics[width=1.\linewidth]{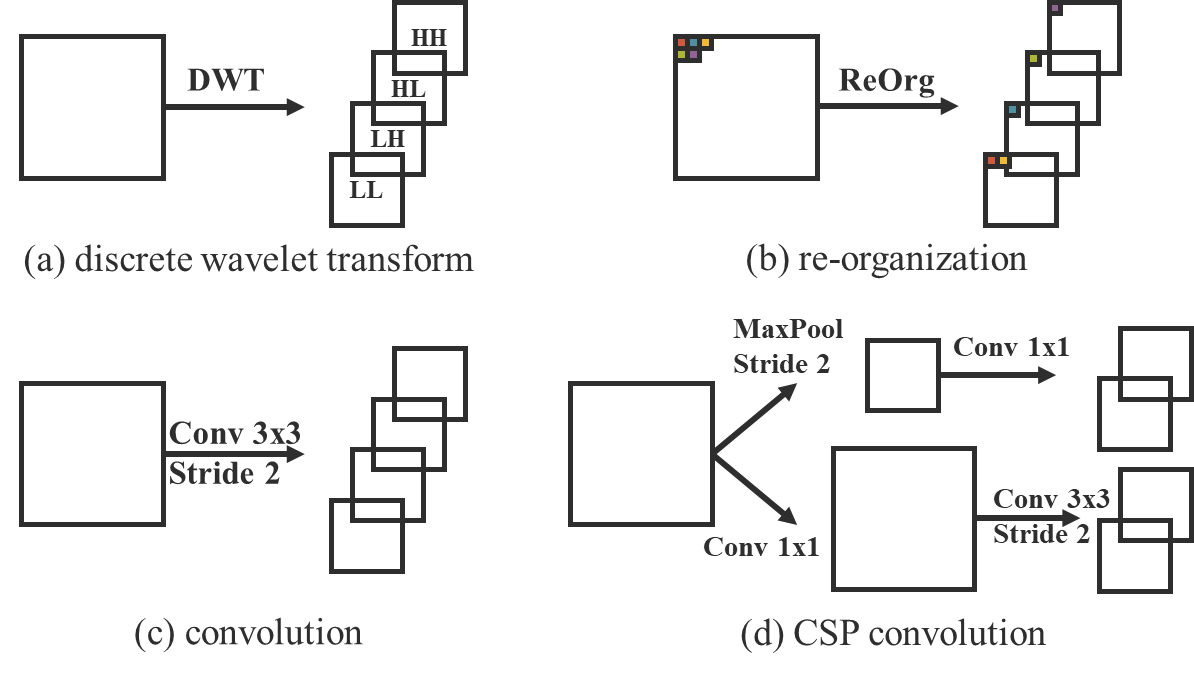}
	\end{center}
	\caption{We use four kind of down-sampling modules in this work, including (a) discrete wavelet transform (DWT): \url{https://github.com/fbcotter/pytorch_wavelets}, (b) re-organization (ReOrg): \url{https://github.com/AlexeyAB/darknet/issues/4662\#issuecomment-608886018}, (c) convolution, and (d) CSP convolution used in CSPNet: \url{https://github.com/WongKinYiu/CrossStagePartialNetworks/tree/pytorch}.}
	\label{fig:down}
\end{figure}

\begin{figure}[h]
\begin{center}
	\includegraphics[width=1.\linewidth]{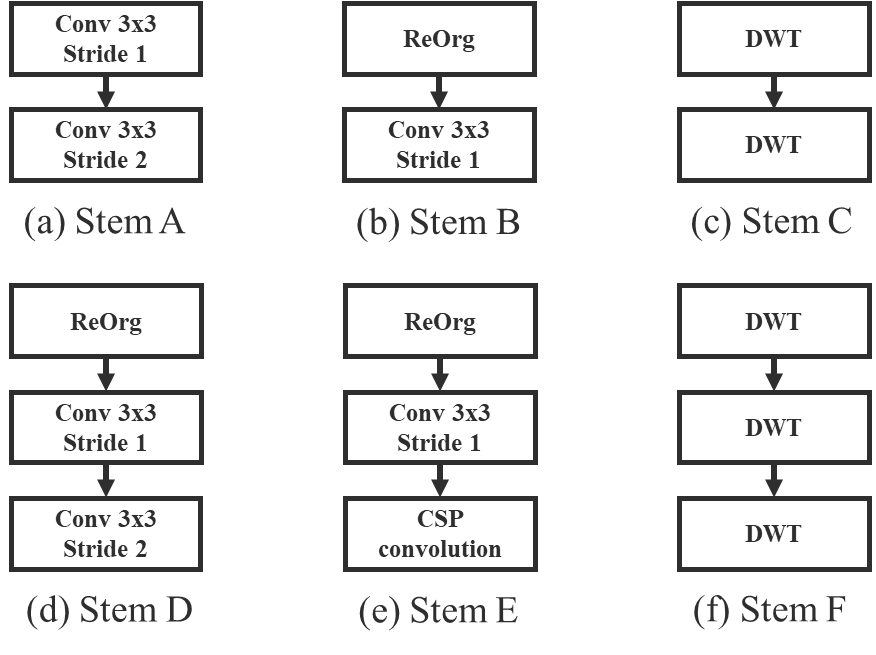}
\end{center}
\caption{We use down-sampling modules in Figure \ref{fig:down} to form stem blocks: (a) Stem A is used in YOLOv4-CSP, (b) Stem B is used in YOLOv4-CSP-fast, (c) Stem C is used in YOLOv4-CSP-SSS, (d) Stem D is proposed by \url{10.5281/zenodo.4679653} and called focus layer, it is used in YOLOv4-P6-light, YOLOR-P6, and YOLOR-W6, (e) Stem E is used in YOLOR-E6 and YOLOR-D6, and (f) Stem F is used in YOLOv4-CSP-SSSS.}
\label{fig:stem}
\end{figure}

\begin{figure}[h]
\begin{center}
	\includegraphics[width=1.\linewidth]{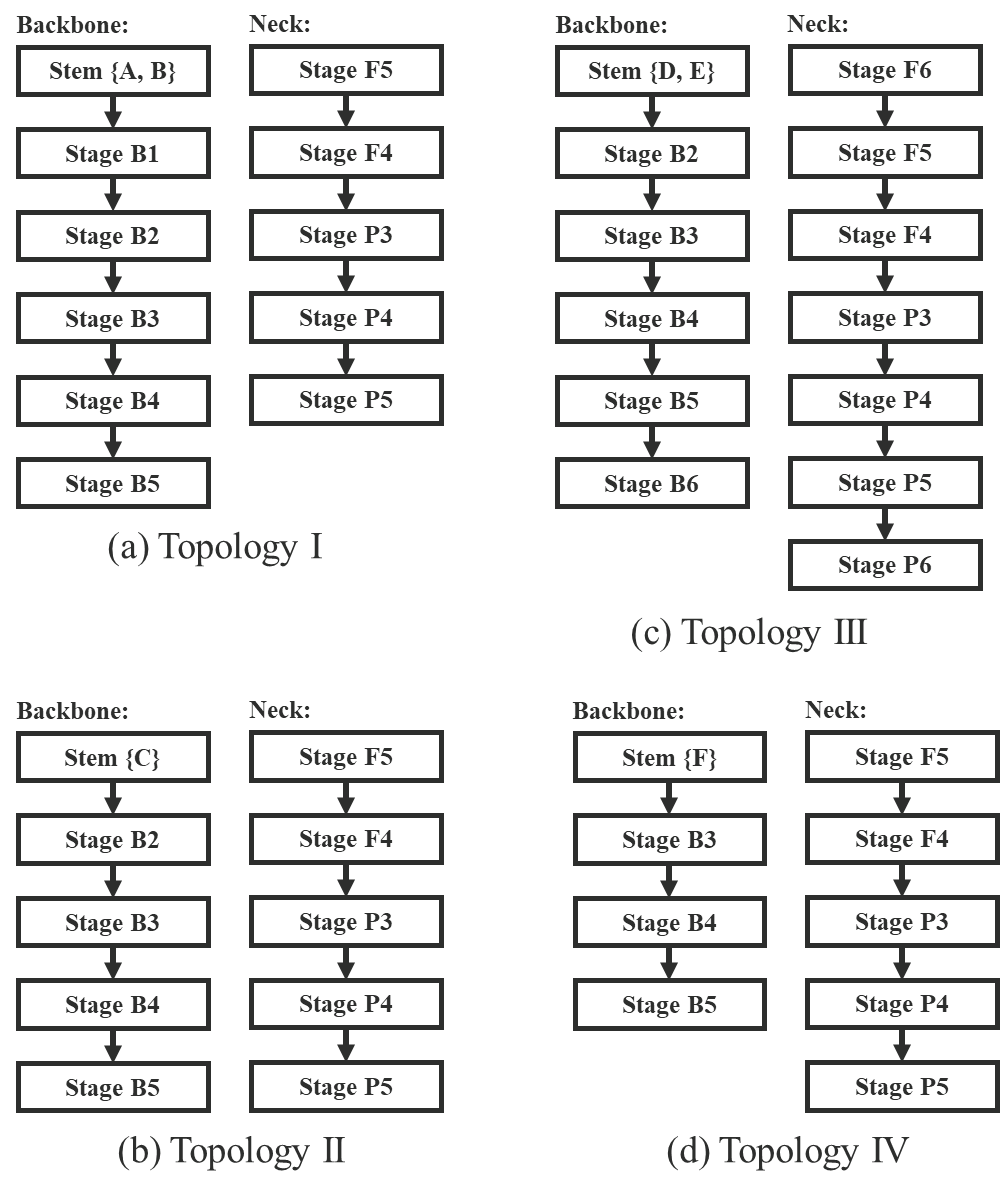}
\end{center}
\caption{Models in this paper can be mapped to three four of architecture topology. Due to Stem C, D, and E contain two down-sampling modules, models used those stem blocks has no Stage B1 in the backbone, for same reason Stem F has no Stage B1 and B2. \\ $\bullet$ YOLOv4-CSP belongs to Topology 1, the architecture is described in Scaled-YOLOv4 paper. \\ $\bullet$ YOLOv4-CSP-fast is modified from YOLOv4-CSP, we replace Stem A in YOLOv4-CSP by Stem B to form YOLOv4-CSP-fast. \\ $\bullet$ YOLOv4-CSP-SSS belongs to Topology 2, Stem C is used in this model. The topology after Stage B2 is as same as YOLOv4-CSP, and width scaling factor and depth scaling factor are set as 0.5 and 0.33, respectively. We then using SiLU activation to replace all Mish activation in the model. \\ $\bullet$ YOLOv4-CSP-SSSS is modified from YOLOv4-CSP-SSS, Stem C in YOLOv4-CSP-SSS is replaced by Stem F in this model. Due to the stem block contains three down-sampling modules, YOLOv4-CSP-SSSS belongs to topology IV. \\ $\bullet$ YOLOv4-P6-light belongs to Topology 3, it uses Stem D and base channels are set as \{128, 256, 384, 512, 640\}. To optimize the gradient propagation, we apply CSP fusion first in B* stages and the repeat number of B2 to B6 are set as \{3, 7, 7, 3, 3\}. \\ $\bullet$ YOLOR-P6 has same architecture as YOLOv4-P6-light, we replace all Mish activation in YOLOv4-P6-light by SiLU activation. \\ $\bullet$ YOLOR-W6 is wider YOLOR-P6, base channels are set as \{128, 256, 512, 768, 1024\}. \\ $\bullet$ YOLOR-E6 expands the width of YOLOR-W6, the width scaling factor is set as 1.25, and all of convolution down-sampling modules are replaced by CSP convolution. \\ $\bullet$ YOLOR-D6 is deeper YOLOR-E6, the repeat number of B2 to B6 are set as \{3, 15, 15, 7, 7\}}
\label{fig:topo}
\end{figure}

\begin{table*}[t]
	\centering
	\begin{threeparttable}[t]
		\footnotesize
		\caption{Lightweight models with implicit knowledge.}
		\label{table:small}
		\setlength\tabcolsep{4.5pt}
		\begin{tabular}{lccccccccccc}
			\toprule
			\textbf{Model} & \textbf{YOLOR} & \textbf{Size} & \textbf{FPS$^{TitanRTX}_{batch 32}$} & \textbf{FLOPs} & \textbf{\# parameters} & \textbf{AP$^{val}$} & \textbf{AP$^{val}_{50}$} & \textbf{AP$^{val}_{75}$} & \textbf{AP$^{val}_{S}$} & \textbf{AP$^{val}_{M}$} & \textbf{AP$^{val}_{L}$} \\				
			\midrule
			\textbf{Y4}-SSS &  & 640 & 720 & 17.9G & 9290077 & 38.8\% & 57.8\% & 42.1\% & 21.3\% & 44.0\% & 52.4\% \\
			\textbf{Y4}-SSS & $\checkmark$ & 640 & 712 & 17.9G & 9291738 & 39.3\% & 58.1\% & 42.5\% & 21.7\% & 44.4\% & 52.8\% \\
			&  &  &  &  & \textbf{\textcolor{red}{+0.018\%}} & \textbf{\textcolor{green}{+0.5\%}} & \textbf{\textcolor{green}{+0.3\%}} & \textbf{\textcolor{green}{+0.4\%}} & \textbf{\textcolor{green}{+0.4\%}} & \textbf{\textcolor{green}{+0.4\%}} & \textbf{\textcolor{green}{+0.4\%}} \\
			\midrule
			\textbf{Y4}-SSSS &  & 640 & 806 & 16.1G & 9205693 & 36.3\% & 54.8\% & 39.3\% & 17.7\% & 40.7\% & 49.7\% \\
			\textbf{Y4}-SSSS & $\checkmark$ & 640 & 791 & 16.1G & 9207354 & 36.8\% & 55.1\% & 39.7\% & 18.9\% & 41.4\% & 50.8\% \\
			&  &  &  &  & \textbf{\textcolor{red}{+0.018\%}} & \textbf{\textcolor{green}{+0.5\%}} & \textbf{\textcolor{green}{+0.3\%}} & \textbf{\textcolor{green}{+0.4\%}} & \textbf{\textcolor{green}{+1.2\%}} & \textbf{\textcolor{green}{+0.7\%}} & \textbf{\textcolor{green}{+1.1\%}} \\
			\midrule
			\textbf{U5R5}-S &  & 640 & 569 & 17.0G & 7266973 & 36.7\% & 55.4\% & 39.8\% & 22.2\% & 41.9\% & 46.2\% \\
			\textbf{U5R5}-S & $\checkmark$ & 640 & 563 & 17.0G & 7268634 & 37.3\% & 56.5\% & 40.5\% & 21.1\% & 42.7\% & 47.7\% \\
			&  &  &  &  & \textbf{\textcolor{red}{+0.023\%}} & \textbf{\textcolor{green}{+0.6\%}} & \textbf{\textcolor{green}{+1.1\%}} & \textbf{\textcolor{green}{+0.7\%}} & \textbf{\textcolor{red}{-1.1\%}} & \textbf{\textcolor{green}{+0.8\%}} & \textbf{\textcolor{green}{+1.5\%}} \\				
			\bottomrule
		\end{tabular}
		\begin{tablenotes}[flushleft]
			\footnotesize
			\item[*] Y4: YOLOv4-CSP, U5R5: \url{10.5281/zenodo.4679653}; FPS: model inference only.
			\item[$\bullet$] \textbf{YOLOR}-Y4-SSSS get 0.1\% better AP than U5R5-S with 39\% faster inference speed.
		\end{tablenotes}
	\end{threeparttable}
\end{table*}

\begin{table*}[t]
	\centering
	\begin{threeparttable}[t]
		\footnotesize
		\caption{Large models with implicit knowledge.}
		\label{table:large}
		\setlength\tabcolsep{5.8pt}
		\begin{tabular}{lcccccccccc}
			\toprule
			\textbf{Model} & \textbf{Size} & \textbf{FPS$^{TitanRTX}_{batch 32}$} & \textbf{FLOPs} & \textbf{\# parameters} & \textbf{AP$^{val}$} & \textbf{AP$^{val}_{50}$} & \textbf{AP$^{val}_{75}$} & \textbf{AP$^{val}_{S}$} & \textbf{AP$^{val}_{M}$} & \textbf{AP$^{val}_{L}$} \\				
			\midrule
			\textbf{YOLOR}-P6 & 1280 & 72 & 326.2G & 37265016 & 52.5\% & 70.6\% & 57.4\% & 37.4\% & 57.3\% & 65.2\% \\
			\textbf{YOLOR}-W6 & 1280 & 66 & 454.0G & 79873400 & 54.0\% & 72.1\% & 59.1\% & 38.1\% & 58.8\% & 67.0\% \\
			\textbf{YOLOR}-E6 & 1280 & 39 & 684.0G & 115909400 & 54.6\% & 72.5\% & 59.8\% & 39.9\% & 59.0\% & 67.9\% \\
			\textbf{YOLOR}-D6 & 1280 & 31 & 936.8G & 151782680 & 55.4\% & 73.5\% & 60.6\% & 40.4\% & 60.1\% & 68.7\% \\
			\midrule
			\textbf{Y4}-P6 & 1280 & 34 & 718.4G & 127530352 & 54.4\% & 72.7\% & 59.5\% & 39.5\% & 58.9\% & 67.3\% \\
			\midrule
			\textbf{U5R5}-S6 & 1280 & 139 & 69.6G & 12653596 & 43.3\% & 61.9\% & 47.7\% & 29.0\% & 48.0\% & 53.3\% \\
			\textbf{U5R5}-M6 & 1280 & 93 & 209.6G & 35889612 & 50.5\% & 68.7\% & 55.2\% & 35.5\% & 55.2\% & 62.0\% \\
			\textbf{U5R5}-L6 & 1280 & 67 & 470.8G & 77218620 & 53.4\% & 71.1\% & 58.3\% & 38.2\% & 58.4\% & 65.7\% \\
			\textbf{U5R5}-X6 & 1280 & 36 & 891.6G & 141755500 & 54.4\% & 72.0\% & 59.1\% & 40.1\% & 59.0\% & 67.2\% \\
			\bottomrule
		\end{tabular}
		\begin{tablenotes}[flushleft]
			\footnotesize
			\item[*] Y4: YOLOv4-CSP, U5R5: \url{10.5281/zenodo.4679653}; FPS: model inference only.
			\item[$\bullet$] Y4-P6 get better AP than U5R5-X6 with 24\% less computation and 11\% fewer \#parameters.
			\item[$\bullet$] \textbf{YOLOR}-E6 get better AP than Y4-P6 with 5\% less computation, 10\% fewer \#parameters, and 15\% faster inference speed.
		\end{tablenotes}
	\end{threeparttable}
\end{table*}

\begin{table*}[t]
\centering
\begin{threeparttable}[t]
	\footnotesize
	\caption{More comparison.}
	\label{table:more}
	\setlength\tabcolsep{5.8pt}
	\begin{tabular}{lcccccccccc}
		\toprule
		\textbf{Model} & \textbf{Size} & \textbf{FPS$^{V/R}$} & \textbf{FLOPs} & \textbf{\# parameters} & \textbf{AP$^{test}$} & \textbf{AP$^{test}_{50}$} & \textbf{AP$^{test}_{75}$} & \textbf{AP$^{test}_{S}$} & \textbf{AP$^{test}_{M}$} & \textbf{AP$^{test}_{L}$} \\				
		\midrule
		\textbf{YOLOR}-P6 & 1280 & 49 / 48  & 326G & 37M & 52.6\% & 70.6\% & 57.6\% & 34.7\% & 56.6\% & 64.2\% \\
		\textbf{YOLOR}-P6D & 1280 & 49 / 48  & 326G & 37M & 53.0\% & 71.0\% & 58.0\% & 35.7\% & 57.0\% & 64.6\% \\
		\textbf{YOLOR}-W6 & 1280 & 47 / 44 & 454G & 80M & 54.1\% & 72.0\% & 59.2\% & 36.3\% & 57.9\% & 66.1\% \\
		\textbf{YOLOR}-E6 & 1280 & 37 / 27 & 684G & 116M & 54.8\% & 72.7\% & 60.0\% & 36.9\% & 58.7\% & 66.9\% \\		
		\textbf{YOLOR}-D6 & 1280 & 30 / 22 & 937G & 152M & 55.4\% & 73.3\% & 60.6\% & 38.0\% & 59.2\% & 67.1\% \\		
		\midrule
		\textbf{ST}-L (HTC++) & -- & -- & 1470G & 284M & 57.7\% & -- & -- & -- & -- & -- \\
		\midrule
		\textbf{C2} (R2) & 1560 & -- / 5 & -- & -- & 56.4\% & 74.0\% & 61.6\% & 38.7\% & 59.7\% & 68.6\% \\
		\midrule
		\textbf{Y4}-P5 & 896 & 41 / -- & 328G & 71M & 51.8\% & 70.3\% & 56.6\% & 33.4\% & 55.7\% & 63.4\% \\
		\textbf{Y4}-P6 & 1280 & 30 / -- & 718G & 128M & 54.5\% & 72.6\% & 59.8\% & 36.8\% & 58.3\% & 65.9\% \\
		\textbf{Y4}-P7 & 1536 & 16 / -- & 1639G & 287M & 55.5\% & 73.4\% & 60.8\% & 38.4\% & 59.4\% & 67.7\% \\
		\midrule
		\textbf{P2} & 640 & 50* / -- & -- & -- & 50.3\% & 69.0\% & 55.3\% & 31.76\% & 53.9\% & 62.4\% \\
		\midrule
		\midrule
		\textbf{Model} & \textbf{Size} & \textbf{FPS$^{V/R}$} & \textbf{FLOPs} & \textbf{\# parameters} & \textbf{AP$^{val}$} & \textbf{AP$^{val}_{50}$} & \textbf{AP$^{val}_{75}$} & \textbf{AP$^{val}_{S}$} & \textbf{AP$^{val}_{M}$} & \textbf{AP$^{val}_{L}$} \\		
		\midrule
		\textbf{ST}-T (MRCNN) & -- & 15.3 / -- & 745G & 86M & 50.5\% & 69.3\% & 54.9\% & -- & -- & -- \\
		\textbf{ST}-S (MRCNN) & -- & 12.0 / -- & 838G & 107M & 51.8\% & 70.4\% & 56.3\% & -- & -- & -- \\
		\textbf{ST}-B (MRCNN) & -- & 11.6 / -- & 982G & 145M & 51.9\% & 70.9\% & 56.5\% & -- & -- & -- \\
		\textbf{ST}-B (HTC++) & -- & -- & 1043G & 160M & 56.4\% & -- & -- & -- & -- & -- \\
		\textbf{ST}-L (HTC++) & -- & -- & 1470G & 284M & 57.1\% & -- & -- & -- & -- & -- \\
		\midrule
		\textbf{C2} (DLA) & 640 & -- / 38 & -- & -- & 49.2\% & -- & -- & -- & -- & -- \\
		\bottomrule
	\end{tabular}
	\begin{tablenotes}[flushleft]
		\footnotesize
		\item[*] ST: SwinTransformer, C2: CenterNet2, Y4: YOLOv4-CSP, P2: PP-YOLOv2.
		\item[*] HTC: Hybrid Task Cascade, R2: Res2Net, MRCNN: Mask R-CNN, DLA: Deep Layer Aggregation.
		\item[*] FPS: end-to-end batch one inference speed of V100/TitanRTX, FPS value with * indicates speed of model inference only.
		\item[$\bullet$] \textbf{YOLOR}-P6D means joint train \textbf{YOLOR}-P6 model with \textbf{YOLOR}-D6 model.
		\item[$\bullet$] \textbf{YOLOR}-D6 get 0.9\% better AP than Y4-P6 with almost same inference speed.
		\item[$\bullet$] \textbf{YOLOR}-D6 get 88\% faster inference speed than Y4-P7 with almost same AP.
	\end{tablenotes}
\end{threeparttable}
\end{table*}


\clearpage
\clearpage
\clearpage

{\small
	\begin{thebibliography}{10}\itemsep=-1pt
	
	\bibitem{aharon2006k}
	Michal Aharon, Michael Elad, and Alfred Bruckstein.
	\newblock {K-SVD}: An algorithm for designing overcomplete dictionaries for
	sparse representation.
	\newblock {\em IEEE Transactions on signal processing}, 54(11):4311--4322,
	2006.
	
	\bibitem{bai2019deep}
	Shaojie Bai, J~Zico Kolter, and Vladlen Koltun.
	\newblock Deep equilibrium models.
	\newblock In {\em Advances in Neural Information Processing Systems (NeurIPS)},
	2019.
	
	\bibitem{bai2020multiscale}
	Shaojie Bai, Vladlen Koltun, and J~Zico Kolter.
	\newblock Multiscale deep equilibrium models.
	\newblock In {\em Advances in Neural Information Processing Systems (NeurIPS)},
	2020.
	
	\bibitem{cao2019gcnet}
	Yue Cao, Jiarui Xu, Stephen Lin, Fangyun Wei, and Han Hu.
	\newblock {GCNet}: Non-local networks meet squeeze-excitation networks and
	beyond.
	\newblock In {\em Proceedings of the IEEE International Conference on Computer
		Vision Workshop (ICCV Workshop)}, 2019.
	
	\bibitem{carion2020end}
	Nicolas Carion, Francisco Massa, Gabriel Synnaeve, Nicolas Usunier, Alexander
	Kirillov, and Sergey Zagoruyko.
	\newblock End-to-end object detection with transformers.
	\newblock In {\em Proceedings of the European Conference on Computer Vision
		(ECCV)}, pages 213--229, 2020.
	
	\bibitem{ghiasi2020simple}
	Golnaz Ghiasi, Yin Cui, Aravind Srinivas, Rui Qian, Tsung-Yi Lin, Ekin~D Cubuk,
	Quoc~V Le, and Barret Zoph.
	\newblock Simple copy-paste is a strong data augmentation method for instance
	segmentation.
	\newblock {\em arXiv preprint arXiv:2012.07177}, 2020.
	
	\bibitem{li2019selective}
	Xiang Li, Wenhai Wang, Xiaolin Hu, and Jian Yang.
	\newblock Selective kernel networks.
	\newblock In {\em Proceedings of the IEEE Conference on Computer Vision and
		Pattern Recognition (CVPR)}, pages 510--519, 2019.
	
	\bibitem{lin2017feature}
	Tsung-Yi Lin, Piotr Doll{\'a}r, Ross Girshick, Kaiming He, Bharath Hariharan,
	and Serge Belongie.
	\newblock Feature pyramid networks for object detection.
	\newblock In {\em Proceedings of the IEEE Conference on Computer Vision and
		Pattern Recognition (CVPR)}, pages 2117--2125, 2017.
	
	\bibitem{lin2014microsoft}
	Tsung-Yi Lin, Michael Maire, Serge Belongie, James Hays, Pietro Perona, Deva
	Ramanan, Piotr Doll{\'a}r, and C~Lawrence Zitnick.
	\newblock {Microsoft COCO}: Common objects in context.
	\newblock In {\em Proceedings of the European Conference on Computer Vision
		(ECCV)}, pages 740--755, 2014.
	
	\bibitem{liu2021swin}
	Ze Liu, Yutong Lin, Yue Cao, Han Hu, Yixuan Wei, Zheng Zhang, Stephen Lin, and
	Baining Guo.
	\newblock Swin transformer: Hierarchical vision transformer using shifted
	windows.
	\newblock {\em arXiv preprint arXiv:2103.14030}, 2021.
	
	\bibitem{sitzmann2020implicit}
	Vincent Sitzmann, Julien Martel, Alexander Bergman, David Lindell, and Gordon
	Wetzstein.
	\newblock Implicit neural representations with periodic activation functions.
	\newblock In {\em Advances in Neural Information Processing Systems (NeurIPS)},
	2020.
	
	\bibitem{sukhbaatar2015end}
	Sainbayar Sukhbaatar, Arthur Szlam, Jason Weston, and Rob Fergus.
	\newblock End-to-end memory networks.
	\newblock In {\em Advances in Neural Information Processing Systems (NeurIPS)},
	2015.
	
	\bibitem{tan2019efficientdet}
	Mingxing Tan, Ruoming Pang, and Quoc~V Le.
	\newblock {EfficientDet}: Scalable and efficient object detection.
	\newblock In {\em Proceedings of the IEEE Conference on Computer Vision and
		Pattern Recognition (CVPR)}, 2020.
	
	\bibitem{vaswani2017attention}
	Ashish Vaswani, Noam Shazeer, Niki Parmar, Jakob Uszkoreit, Llion Jones,
	Aidan~N Gomez, Lukasz Kaiser, and Illia Polosukhin.
	\newblock Attention is all you need.
	\newblock In {\em Advances in Neural Information Processing Systems (NeurIPS)},
	2017.
	
	\bibitem{wang2020scaled}
	Chien-Yao Wang, Alexey Bochkovskiy, and Hong-Yuan~Mark Liao.
	\newblock {Scaled-YOLOv4}: Scaling cross stage partial network.
	\newblock {\em Proceedings of the IEEE Conference on Computer Vision and
		Pattern Recognition (CVPR)}, 2021.
	
	\bibitem{wang2016robust}
	Chien-Yao Wang, Seksan Mathulaprangsan, Bo-Wei Chen, Yu-Hao Chin, Jing-Jia
	Shiu, Yu-San Lin, and Jia-Ching Wang.
	\newblock Robust face verification via bayesian sparse representation.
	\newblock In {\em 2016 Asia-Pacific Signal and Information Processing
		Association Annual Summit and Conference (APSIPA)}, pages 1--4, 2016.
	
	\bibitem{wang2017recognition}
	Chien-Yao Wang, Andri Santoso, Seksan Mathulaprangsan, Chin-Chin Chiang,
	Chung-Hsien Wu, and Jia-Ching Wang.
	\newblock Recognition and retrieval of sound events using sparse coding
	convolutional neural network.
	\newblock In {\em 2017 IEEE International Conference on Multimedia and Expo
		(ICME)}, pages 589--594, 2017.
	
	\bibitem{wang2020sound}
	Chien-Yao Wang, Tzu-Chiang Tai, Jia-Ching Wang, Andri Santoso, Seksan
	Mathulaprangsan, Chin-Chin Chiang, and Chung-Hsien Wu.
	\newblock Sound events recognition and retrieval using
	multi-convolutional-channel sparse coding convolutional neural networks.
	\newblock {\em IEEE/ACM Transactions on Audio, Speech, and Language Processing
		(TASLP)}, 28:1875--1887, 2020.
	
	\bibitem{wang2020implicit}
	Tiancai Wang, Xiangyu Zhang, and Jian Sun.
	\newblock Implicit feature pyramid network for object detection.
	\newblock {\em arXiv preprint arXiv:2012.13563}, 2020.
	
	\bibitem{wang2021pyramid}
	Wenhai Wang, Enze Xie, Xiang Li, Deng-Ping Fan, Kaitao Song, Ding Liang, Tong
	Lu, Ping Luo, and Ling Shao.
	\newblock Pyramid vision transformer: A versatile backbone for dense prediction
	without convolutions.
	\newblock {\em arXiv preprint arXiv:2102.12122}, 2021.
	
	\bibitem{wang2018non}
	Xiaolong Wang, Ross Girshick, Abhinav Gupta, and Kaiming He.
	\newblock Non-local neural networks.
	\newblock In {\em Proceedings of the IEEE Conference on Computer Vision and
		Pattern Recognition (CVPR)}, pages 7794--7803, 2018.
	
	\bibitem{weston2014memory}
	Jason Weston, Sumit Chopra, and Antoine Bordes.
	\newblock Memory networks.
	\newblock In {\em International Conference on Learning Representations (ICLR)},
	2015.
	
	\bibitem{wright2008robust}
	John Wright, Allen~Y Yang, Arvind Ganesh, S~Shankar Sastry, and Yi Ma.
	\newblock Robust face recognition via sparse representation.
	\newblock {\em IEEE Transactions on Pattern Analysis and Machine Intelligence
		(TPAMI)}, 31(2):210--227, 2008.
	
	\bibitem{yin2020disentangled}
	Minghao Yin, Zhuliang Yao, Yue Cao, Xiu Li, Zheng Zhang, Stephen Lin, and Han
	Hu.
	\newblock Disentangled non-local neural networks.
	\newblock In {\em Proceedings of the European Conference on Computer Vision
		(ECCV)}, pages 191--207, 2020.
	
	\bibitem{zhang2020resnest}
	Hang Zhang, Chongruo Wu, Zhongyue Zhang, Yi Zhu, Zhi Zhang, Haibin Lin, Yue
	Sun, Tong He, Jonas Mueller, R Manmatha, et~al.
	\newblock {ResNeSt}: Split-attention networks.
	\newblock {\em arXiv preprint arXiv:2004.08955}, 2020.
	
	\bibitem{zhou2021probabilistic}
	Xingyi Zhou, Vladlen Koltun, and Philipp Kr{\"a}henb{\"u}hl.
	\newblock Probabilistic two-stage detection.
	\newblock {\em arXiv preprint arXiv:2103.07461}, 2021.
	
	\end{thebibliography}
}
\end{document}